\documentclass{article} 
\usepackage{iclr2027_conference,times}

\usepackage{amsmath,amsfonts,bm}

\def\eqref#1{equation~\ref{#1}}

\def\1{\bm{1}}

\DeclareMathAlphabet{\mathsfit}{\encodingdefault}{\sfdefault}{m}{sl}
\SetMathAlphabet{\mathsfit}{bold}{\encodingdefault}{\sfdefault}{bx}{n}

\usepackage{hyperref}
\usepackage{url}

\usepackage{graphicx}
\usepackage{svg}
\usepackage{booktabs}
\usepackage{pifont}
\usepackage{multirow}
\usepackage{tikz}
\usetikzlibrary{positioning,arrows.meta,fit,backgrounds}

\usepackage{wrapfig}
\usepackage{enumitem}
\usepackage{caption}
\newcommand{\cmark}{\textcolor{green!60!black}{\ding{51}}}
\newcommand{\xmark}{\textcolor{red!75!black}{\ding{55}}}
\newcommand{\std}[1]{{\scriptscriptstyle\,\pm #1}}

\newcommand{\Epos}{E_{\mathrm{pos}}}

\newcommand{\Eori}{E_{\mathrm{ori}}}

\newcommand{\Evel}{E_{\mathrm{vel}}}

\definecolor{mycolor}{RGB}{254,113,52}
\title{\mbox{\textcolor{mycolor}{HIPC}: \textcolor{mycolor}{H}umanoid \textcolor{mycolor}{I}terative \textcolor{mycolor}{P}lanning and \textcolor{mycolor}{C}on-}\\
\mbox{trol via Goal-Conditioned Motion Closure}}
\author{
\begin{tabular}{@{}l@{}}
    \normalfont
    Yi Han\textsuperscript{1}\thanks{Equal Contribution. $^\dagger$Equal Advising.},\ 
    Xingyu Chen\textsuperscript{2,3}$^*$,\ 
    Zixuan Lang\textsuperscript{3},\ 
    Qianyang Wu\textsuperscript{4},\ 
    Yuhang Lin\textsuperscript{5},\ 
    Ji Ma\textsuperscript{6},\ 
    Liang Li\textsuperscript{7},\\
    \normalfont
    Dongting Li\textsuperscript{7},\ 
    Xinpeng Liu\textsuperscript{1},\ 
    Ruoli Dai\textsuperscript{3},\ 
    Hesheng Wang\textsuperscript{1}$^\dagger$,\ 
    Renjing Xu\textsuperscript{2}$^\dagger$,\ 
    Lei Han\textsuperscript{3}$^\dagger$\\
    \normalfont\mdseries\small
    \textsuperscript{1}SJTU \quad
    \textsuperscript{2}HKUST(GZ) \quad
    \textsuperscript{3}Noitom \quad
    \textsuperscript{4}ShanghaiTech \quad
    \textsuperscript{5}ZJU \quad
    \textsuperscript{6}HKU \quad
    \textsuperscript{7}THU
\end{tabular}\\
\centerline{\href{https://hipc-humanoid.github.io/}
{\texttt{https://hipc-humanoid.github.io/}}}
\vspace{-25pt}
}
\iclrfinalcopy 
\begin{document}

\maketitle
\begin{figure}[h]
    \centering
    \includegraphics[width=1.0\textwidth]{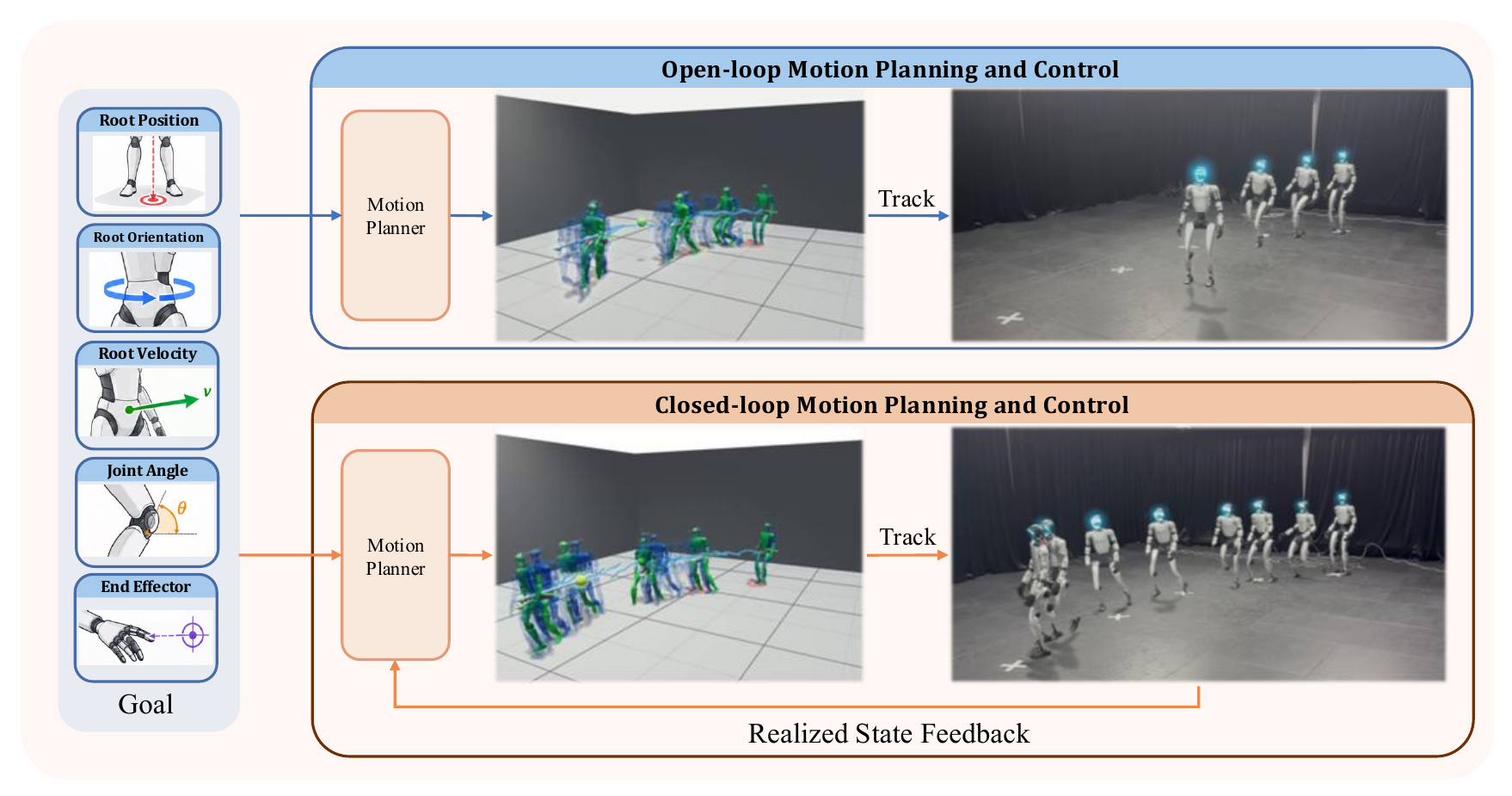}
    \caption{
We propose Humanoid Interactive Planning and Control (HIPC), a real-time asynchronous closed-loop framework for goal-reaching task. HIPC feeds back realized robot states into the planner, enabling the system to interpret and execute multimodal conditions, including root pose, joint angle, end-effector, natural-language instructions, and their combinations, and continuously correct accumulated spatial drift.
    }
    \label{fig:teaser}
\end{figure}
\begin{abstract}
Recent generative models are widely used to provide whole-body motion for humanoid tracking. Current methods combine motion generators with controllers in an open-loop paradigm, which leads to critical issues: tracking errors can accumulate into spatial drift, and the planner cannot adapt to unexpected physical disturbances without feedback from physical execution. 
However, directly feeding realized states back to the planner in a closed loop can cause discontinuities, resulting in physical jitter and even falls, since such states are out of distribution during training.
To bridge this gap, we propose \textbf{HIPC}, a real-time \textbf{Humanoid Iterative Planning and Control} framework that couples a generative model with motion tracking into a unified planning-and-control paradigm.
HIPC introduces a goal-conditioned motion planning trained with a canonical motion representation that avoids gimbal-lock singularities under extreme orientations and supports motion generation on diverse geometric and semantic conditions.
Then, HIPC adopts two strategies, deviation-decayed recovery and history re-anchoring, to mitigate the out-of-distribution shift caused by realized-state feedback.
Finally, HIPC operates a real-time asynchronous planning-and-control system, with latency compensation to reduce temporal mismatch for inference delay.
Extensive experiments confirm the effectiveness of HIPC under switching and composition of diverse conditions. Results show that HIPC achieves state-of-the-art goal-reaching accuracy among existing methods while simultaneously demonstrating robustness to physical perturbations and the ability to recover from falls.
\end{abstract}

\newpage
\section{Introduction}


Recent general motion trackers~\citep{luo2026sonic,zeng2026scaling,mimiclite2026} have made diverse whole-body behaviors executable through a single policy, providing a reusable control interface and shifting the challenge upstream from task-specific control to motion planning.
Early works like optimization-based planning~\citep{kuindersma2016optimization} is difficult to scale up due to high-dimensional whole-body coordination and complex contact constraints~\citep{gu2026evolution}, motivating generative planners that synthesize diverse references for general tracker~\citep{xie2026textop}.
Although the recent planner~\citep{chen2026reactivebfm} provides explicit objectives that guide whole-body movements to goal, it still lacks diverse goal condition type and precise closed-loop execution.

Composing a generative planner with a whole-body tracker leaves the planning-to-control interface open-loop~\citep{kaynov2008open}, Since physical execution inevitably deviates from kinematic references, creating a \emph{Tracking Gap}.
Although the tracker locally corrects such deviations, the tracking errors remain invisible to the planner and can accumulate into spatial drift, motivating motion planning control paradigm to maintain spatial consistency.
Meanwhile, physical disturbances and imperfect state estimation may cause the robot state to deviate from the planned motion~\citep{rotella2014state}, potentially exceeding the tracker's recovery capability and requiring the planner to adapt motion generation through body-state feedback.
Both challenges call for incorporating realized physical states into generative motion planning. 
Nevertheless, directly incorporating realized states causes an \emph{Execution Distribution Gap}, as physical execution histories differ from the kinematic distribution during training.
Existing self-rollout strategies mitigate autoregressive distribution shift~\citep{chen2026reactivebfm}, but do not directly capture structured deviations induced by downstream dynamics.

Moreover, replanning from realized states that deviate from the current reference can create discontinuities between successive motion chunks. Non-negligible \emph{Inference Delay}~\citep{black2026real,black2025training} further misaligns planning and execution as the robot continues to evolve during generation.
Together, these effects can produce abrupt reference transitions, physical jitter, and even loss of balance.
Closed-loop motion-data refinement can incorporate controller feedback during training~\citep{chen2026human}, but couples motion synthesis to a specific downstream controller~\citep{xu2026iterative}.
The key challenge is to close the planning-control loop while enabling smooth physical execution without sacrificing planner-controller modularity.

In this paper, we introduce \textbf{HIPC}, a real-time \textbf{Humanoid Iterative Planning and Control} framework that closes the loop between generative motion planning and whole-body tracking.
HIPC combines a more diverse goal-conditioned motion planner with execution-consistent state feedback and latency-aware asynchronous architecture.
It corrects accumulated spatial drift and supports recovery from physical disturbances while preserving planner-native motion histories.
It further anticipates the state reached during planner inference for smooth reference handoff.
This enables closed-loop whole-body planning without requiring controller-specific rollout data or sacrificing planner--controller modularity. 

In summary, the primary contributions of this work are three-fold:

\begin{itemize}[leftmargin=0pt, label={}, labelsep=5pt, itemindent=0pt, parsep=0pt]

\item \textbf{Goal-Conditioned Motion Planner with Canonical Representation.}
We introduce a generative model in a goal reaching paradigm as motion planner, and train this generator with a canonical motion representation conditioned on diverse geometric goals and semantic text.

\item \textbf{Execution-Consistent Motion Closure for Real State Feedback.}
We propose a history re-anchoring strategy to incorporate realized robot states into the planning history while keeping it close to the planner's training distribution.

\item \textbf{Latency-Aware Asynchronous Planning and Control.}
We decouple low-frequency generative planning from high-frequency whole-body tracking through a non-blocking asynchronous dual-system architecture, which maintains continuous real-time control and smooth reference handoff during close-loop replanning.

\end{itemize}

\section{Related Work}
\begin{table*}[t]
\centering
\caption{
\textbf{Functional comparison of representative humanoid motion generation
and planning--control approaches.}
``Real-State Feedback'' denotes feedback of realized physical states to the
planner for online replanning.
``Joint'' denotes joint-level configuration goals.
``Perturbation'' and ``Get Up'' denote recovery from external perturbations
and fallen states, respectively.
}

\label{tab:related_work_comparison}

\setlength{\tabcolsep}{4.0pt}
\renewcommand{\arraystretch}{1.15}
\small

\resizebox{\textwidth}{!}{%
\begin{tabular}{@{}lccc|cccc|cc@{}}
\toprule

\multirow{2}{*}{\textbf{Method}}
&
\multirow{2}{*}{\textbf{Text}}
&
\multirow{2}{*}{\textbf{Real-Time}}
&
\multirow{2}{*}{\shortstack{\textbf{Real-State}\\\textbf{Feedback}}}
&
\multicolumn{4}{c|}{\textbf{Goal Condition}}
&
\multicolumn{2}{c}{\textbf{OOD Recovery}}
\\

\cmidrule(lr){5-8}
\cmidrule(lr){9-10}

&
&
&
&
\textbf{Root}
&
\textbf{End Effector}
&
\textbf{Velocity}
&
\textbf{Joint}
&
\textbf{Perturbation}
&
\textbf{Get Up}
\\

\midrule

Kimodo
& \cmark
& \xmark
& \xmark
& \cmark
& \cmark
& \xmark
& \cmark
& \xmark
& \xmark
\\

TextOp
& \cmark
& \cmark
& \xmark
& \xmark
& \xmark
& \xmark
& \xmark
& \xmark
& \xmark
\\

ARDY
& \cmark
& \cmark
& \xmark
& \cmark
& \cmark
& \cmark
& \cmark
& \xmark
& \xmark
\\

OMG
& \cmark
& \cmark
& \xmark
& \xmark
& \xmark
& \xmark
& \xmark
& \xmark
& \xmark
\\

ReactiveBFM
& \cmark
& \cmark
& \cmark
& \cmark
& \xmark
& \xmark
& \xmark
& \cmark
& \cmark
\\

\midrule

\textbf{HIPC (Ours)}
& \cmark
& \cmark
& \cmark
& \cmark
& \cmark
& \cmark
& \cmark
& \cmark
& \cmark
\\

\bottomrule
\end{tabular}%
}
\vspace{-7pt}
\end{table*}

\paragraph{Human Motion Generation.}
Human motion generation has largely evolved along two complementary paradigms. Diffusion-based models such as MDM~\citep{tevet2023human} model continuous motion distributions through iterative denoising, while autoregressive approaches such as MotionGPT~\citep{jiang2023motiongpt} formulate motion generation as sequential prediction in a unified motion--language space. Recent methods increasingly combine the strengths of advanced architectures through masked modeling~\citep{guo2024momask, chen2024text}, hybrid conditions~\citep{zhao2025dartcontrol,li2025genmo}, and scaling controllable modeling~\citep{rempe2026kimodo,zhao2026autoregressive}. HIPC adopts the autoregressive diffusion paradigm as its motion planner backbone.

\paragraph{Humanoid Whole-Body Control.}
Physics-based motion imitation established a powerful framework for acquiring dynamic motor skills from reference motion. Early methods such as DeepMimic~\citep{peng2018deepmimic} and AMP~\citep{peng2021amp} mainly focus on tracking individual motions or limited skill sets, while later systems such as HumanX~\citep{wang2026humanx}, HAIC~\citep{li2026haic}, and VAIC~\citep{li2026vaic} demonstrate increasingly agile whole-body behaviors and dynamic interactions. Recent works further scale motion data and model capacity to build universal motion trackers, such as SONIC~\citep{luo2026sonic} and ScaleBFM~\citep{zeng2026scaling}, consolidating diverse behaviors into reusable low-level control policies. HIPC builds on this progress by coupling its generative planner with a general-purpose whole-body tracker as the execution interface.

\paragraph{Humanoid Generative Motion Control.}
Generative control first emerged in humanoid robotics as a motion prior, GMP~\citep{zhang2025natural}, PDF-HR~\citep{gu2026hr} and PFM-HR~\citep{gao2026pfm} learn a motion distribution to improve policy exploration. Meanwhile, in physics-based character animation CLoSD~\citep{tevet2025closd} couples an autoregressive diffusion planner with an RL tracker and closes the loop through simulated physical execution.
Recent works extend this paradigm to physical humanoids. 
MotionBricks~\citep{wang2026motionbricksscalablerealtimemotions} deploys a motion generator on a humanoid, but leaves the execution-side coupling problem unexplored.
Heracles~\citep{tao2026heracles} identifies large state deviations as a key limitation of rigid reference tracking and introduces a state-conditioned diffusion middleware that preserves tracking near the reference while generating recovery motions under large deviations.
TextOp~\citep{xie2026textop} demonstrates real-time text-driven motion generation and execution, while ReactiveBFM~\citep{chen2026reactivebfm} further studies execution feedback and asynchronous replanning.
HIPC addresses the resulting execution-history and temporal-alignment issues while preserving a modular planner-controller interface.
\section{Proposed Method}
\begin{figure}[ht]
    \centering
    \includegraphics[width=1.0\textwidth]{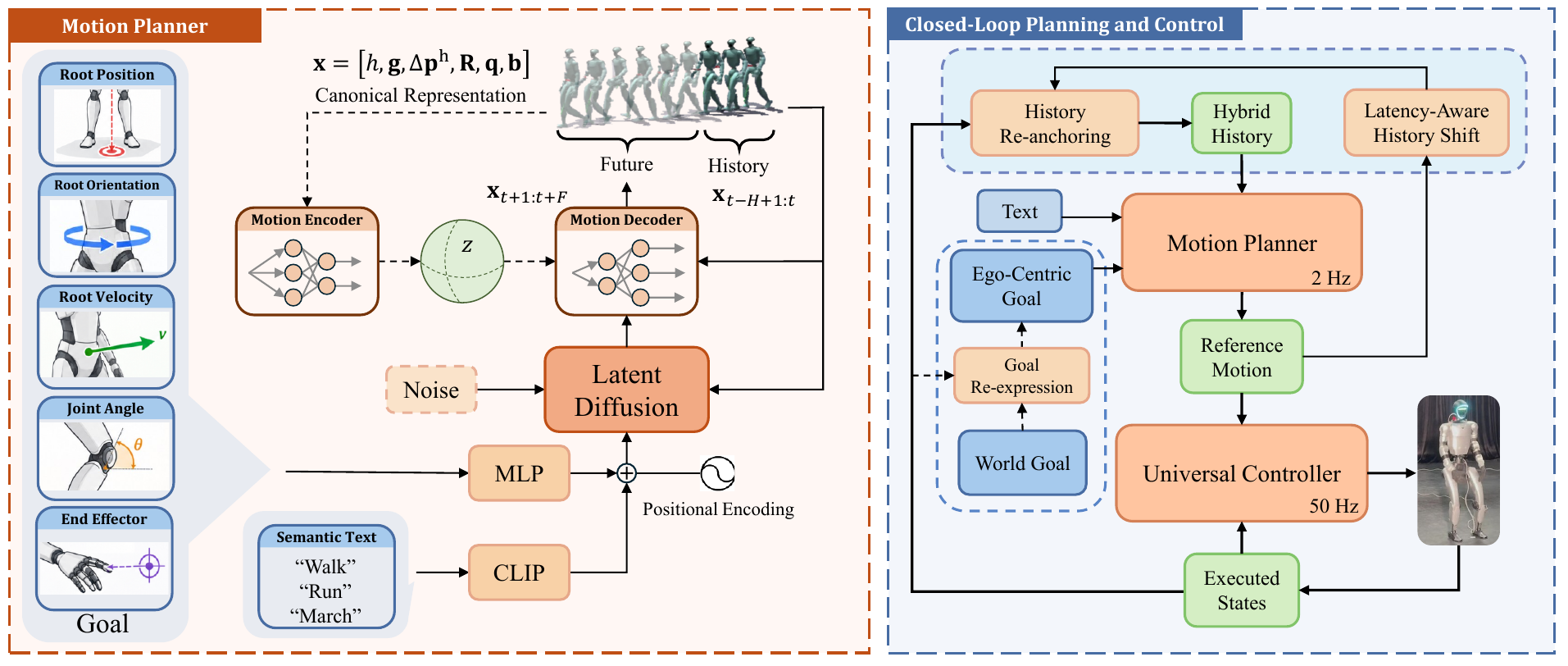}
    \caption{
        Interactive text-conditioned Generated Motion Sequence on motion controller.
    }
    \label{fig:pipeline}
    \vspace{-8pt}
\end{figure}

In this section, we present our closed-loop humanoid planning-and-control framework HIPC, as illustrated in Fig.~\ref{fig:pipeline}. We first introduce canonical motion representation of the goal-conditioned motion planner in Sec.~\ref{sec:goal_planner}. Next, Sec.~\ref{sec:motion_closure} addresses the distribution shift caused by directly feeding realized physical states back into the planner through execution-aware data preparation and history reanchoring. Finally, Sec.~\ref{sec:async_pnc} presents the asynchronous planning-and-control system used for real-world deployment, together with latency compensation for continuous real-time replanning.

\subsection{Goal-Conditioned Motion Planning}
\label{sec:goal_planner}
We formulate whole-body planning as conditional autoregressive motion generation.
At each replanning step $t$, the planner observes the most recent $H=16$ motion frames
$\mathbf{x}_{t-H+1:t}$ together with an optional condition set $\mathbf{c}_t$, and predicts a future motion chunk of $F=64$ frames:
\begin{equation}
\hat{\mathbf{x}}_{t+1:t+F}
=
G_\theta
\left(
\mathbf{x}_{t-H+1:t},
\mathbf{c}_t
\right).
\label{eq:planner_formulation}
\end{equation}
The generated chunk serves as a kinematic reference for the downstream universal whole-body controller.
The condition set $\mathbf{c}_t$ supports both geometric goals and semantic text commands, enabling diverse goal-directed whole-body motion generation.

\paragraph{Canonical Motion Representation.}
Existing motion representations commonly decompose root orientation into roll, pitch, and yaw and construct a heading-aligned local frame, which can suffer from gimbal-lock singularities under large body rotations. We instead represent root motion through relative rigid-body transformations between consecutive root frames and treat horizontal and vertical quantities asymmetrically, using incremental horizontal quantities and direct vertical states. This canonical representation is invariant to global horizontal translation and heading and remains continuous across arbitrary root orientations, including upright, tilted, and fallen configurations. We define each motion frame as
\begin{equation}
\mathbf{x}_t
=
\left[
h_t,
\mathbf{g}_t,
{}^{E_{t-1}}\Delta\mathbf{p}_{t-1}^{\mathrm{h}},
{}^{E_{t-1}}\mathbf{R}_{E_t},
\mathbf{q}_t,
\mathbf{b}_t
\right],
\label{eq:motion_feature}
\end{equation}
where $\mathbf{R}_t\in\mathrm{SO}(3)$ and ${}^{W}\mathbf{p}_t\in\mathbb{R}^3$ denote the root orientation and position in the world frame, respectively. Here, $h_t$ is the root height above ground, $\mathbf{g}_t=\mathbf{R}_t^\top\mathbf{g}\in\mathbb{R}^{3}$ is the gravity direction expressed in the current root frame, $\mathbf{q}_t\in\mathbb{R}^{29}$ contains the joint angles of the 29-DoF Unitree G1, and $\mathbf{b}_t\in\{0,1\}^{2}$ denotes binary foot-contact states. The relative root translation and rotation are defined as ${}^{E_{t-1}}\Delta\mathbf{p}_{t-1}=\mathbf{R}_{t-1}^{\top}({}^W\mathbf{p}_{t}-{}^W\mathbf{p}_{t-1})$ and ${}^{E_{t-1}}\mathbf{R}_{E_t}=\mathbf{R}_{t-1}^{\top}\mathbf{R}_t$, respectively, with ${}^{E_{t-1}}\Delta\mathbf{p}_{t-1}^{\mathrm{h}}$ obtained by projecting ${}^{E_{t-1}}\Delta\mathbf{p}_{t-1}$ onto the horizontal plane orthogonal to $\mathbf{g}_{t-1}$. All rotations provided to the network are encoded using the continuous 6D representation~\citep{zhou2019continuity}, yielding a 44-dimensional motion representation.

\paragraph{Goal-Conditioned Motion Generation.}
The planner supports optional geometric and semantic conditions, which can be provided independently, jointly, or omitted entirely.
Geometric conditions specify explicit goals at time $t^\star$, including root position ${}^{E_t}\Delta\mathbf{p}_{t^\star}^{\mathrm{root}}$, root orientation ${}^{E_t}\mathbf{R}_{t^\star}^{\mathrm{root}}$, root linear velocity ${}^{E_t}\mathbf{v}_{t^\star}^{\mathrm{root}}$, joint configuration $\mathbf{q}_{t^\star}$, and hand and foot end-effector positions ${}^{E_t}\Delta\mathbf{p}_{t^\star}^{\mathrm{ee}}\in\mathbb{R}^{4\times3}$.
An optional time-to-arrival condition resolves temporal ambiguity by specifying when the goal should be reached.
At each replanning step, spatial goals are re-expressed in the current root coordinate system, allowing the planner to respond naturally to changing targets despite being trained on fixed motion clips.
Text provides complementary semantic guidance for motion styles that are difficult to specify through sparse geometric goals, rather than encoding complex task procedures.
We show in Sec.~\ref{sec:experiments} that this formulation supports diverse static and dynamic goal-conditioned applications.

Our planner adopts the VAE--LDM formulation used in DART~\citep{zhao2025dartcontrol} and TextOp~\citep{xie2026textop}.
We first train a VAE to encode future motion chunks into a compact latent space and reconstruct them conditioned on recent motion history, and then freeze the VAE while training a latent diffusion model over the motion latents.
During LDM training, we randomly sample a target frame from each future chunk, use its geometric state to construct the goal condition, and take the corresponding frame offset as the time-to-arrival condition.
The sampled goal frame also provides the geometric goal error used for auxiliary supervision.
To reduce autoregressive exposure bias, we further adopt self-rollout training, progressively replacing ground-truth history with model-generated motion during sequential chunk prediction.

\subsection{Execution-Consistent Motion Closure}
\label{sec:motion_closure}
Closing the motion loop requires the planner to remain reliable across diverse geometric goals, absorb realized-state feedback without leaving the training distribution, and recover from execution-induced deviations.
We address these challenges through goal-aware data construction, state-history fusion, and recovery-oriented training.

\paragraph{Motion Clips with Geometry Coverage.}
\begin{figure}[t]
    \centering
    \includegraphics[width=1.0\textwidth]{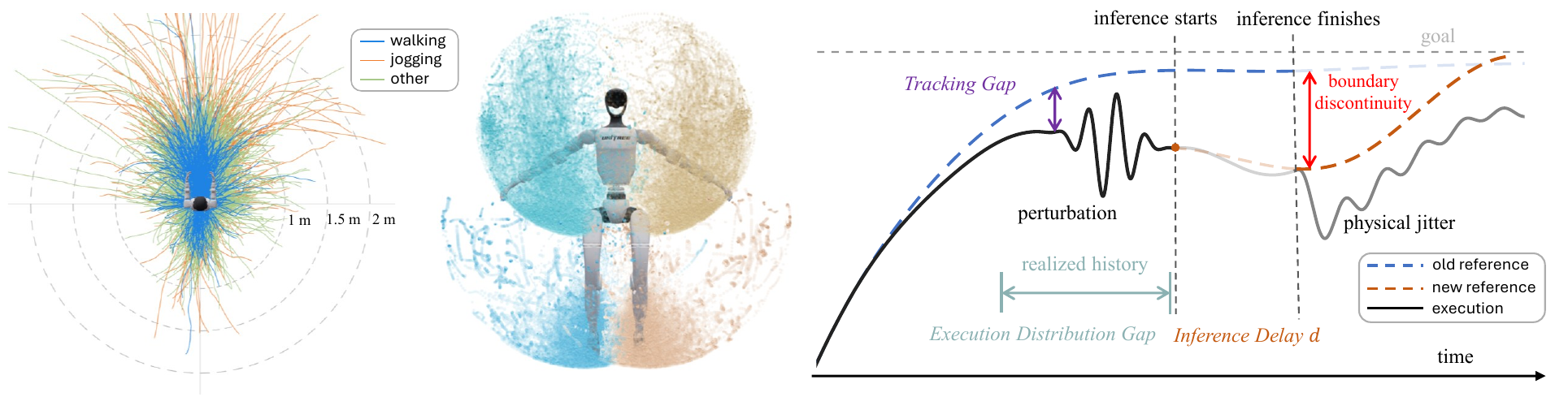}
    \caption{
        \textbf{Geometry coverage and planning--control gaps.}
        (a) Egocentric root-trajectory distribution in the training data.
        (b) Data-derived reachable workspace of hand and foot end-effectors.
        (c) Planning--control gaps caused by tracking deviation, execution-distribution shift, and inference delay, leading to reference discontinuities and physical jitter.
    }
    \label{fig:dataproc}
    \vspace{-8pt}
\end{figure}

To make training motion cover the geometry goal as much space as possible, we use retargeted AMASS~\citep{mahmood2019amass}, LAFAN~\citep{harvey2020robust} and HiPHI~\citep{ji2026hiphi} through GMR~\citep{araujo2025retargeting}, segmented into fixed-length clips, and converted to the canonical representation in Eq.~(\ref{eq:motion_feature}).
We geometrically filter the retargeted motions by removing clips with severe self-collision or sustained center-of-mass displacement far outside the support region of the feet, and further discard motion categories that do not admit meaningful geometric goal specification.
After filtering, we obtain 33,488 training clips and 1,764 validation clips across 189 action categories.
To further improve directional coverage of geometric goals, we increase the sampling probability of clips associated with less frequent target directions during training. The resulting egocentric root and end-effector distributions are shown in Fig.~\ref{fig:dataproc}(a,b), covering a broad range of reachable goals for real-world deployment.


\paragraph{Hybrid State via History Re-anchoring.}
At planning time $t$, the realized robot state generally deviates from its planner prediction. We construct each hybrid history state by applying the current tracking deviation consistently to the generated history:
\begin{equation}
\mathbf{s}_{\tau}^{\mathrm{hyb}}
=
\boldsymbol{\delta}_t
\oplus
\mathbf{s}_{\tau}^{\mathrm{pred}},
\qquad
\boldsymbol{\delta}_t
=
\mathbf{s}_{t}^{\mathrm{real}}
\ominus
\mathbf{s}_{t}^{\mathrm{pred}},
\label{eq:history_reanchoring}
\end{equation}
where $\tau$ denotes an arbitrary frame in the planner history. The operators $\ominus$ and $\oplus$ denote state-wise deviation and composition, respectively, using rigid-body composition for the root pose and additive offsets for joint configurations. In particular, $\mathbf{s}_{t}^{\mathrm{hyb}}=\mathbf{s}_{t}^{\mathrm{real}}$, so the latest realized state is directly incorporated into the planner history, and earlier frames are reconstructed from planner-generated kinematics rather than controller-executed trajectories, avoiding dynamics-dependent execution histories and their high-frequency fluctuations. The reanchored states are then converted to the canonical motion representation $\mathbf{x}$ and used as planner history.

\paragraph{Deviation-Decayed Recovery.}
To expose the planner to tracking deviations encountered by the hybrid history at deployment, we apply the same uniformly sampled deviation $\tilde{\boldsymbol{\delta}}$ coherently to the entire history using the state transformation in Eq.~(\ref{eq:history_reanchoring}).
Over the future trajectory, the deviation progressively decays within a random recovery horizon, guiding the motion back toward the original ground-truth trajectory.
This constructs a recoverable neighborhood around the motion manifold, allowing the planner to correct tracking deviations before they accumulate into persistent planning drift. For get-up motions, we apply stronger deviations only to the preceding fallen states while preserving the original get-up trajectory. This constructs additional recovery samples from diverse fallen configurations that converge to plausible pre-get-up states before entering the original get-up motion, increasing both the diversity and proportion of get-up examples in training.

\subsection{Asynchronous Planning and Control Mechanism}
\label{sec:async_pnc}
\begin{figure}[t]
    \centering
    \includegraphics[width=1.0\textwidth]{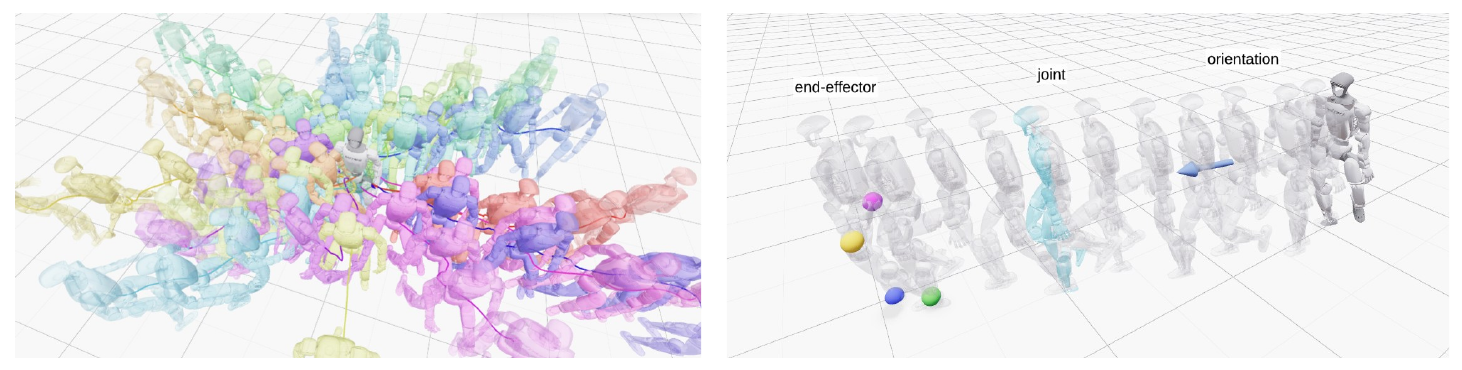}
    \caption{
        \textbf{Static and switching goal reaching.}
        Left: Trajectories from diverse initial states converge to a shared fixed goal.
        Right: The robot continuously adapts its motion to switching goals of different types.
    }
    \label{fig:goal_attraction}
    \vspace{-5pt}
\end{figure}

For real-time deployment, we propose two mechanisms for asynchronous planning and control via a fast-in-slow dual-system architecture and latency-aware history shifting. Asynchronous system try to keep high-frequency control non-blocking and reducing reference discontinuity caused by inference latency. Although asynchronous execution prevents planner inference from blocking the controller, the robot continues following the previous reference during generation and may therefore advance beyond the state from which the new chunk was planned.

\paragraph{Fast-in-Slow Dual-System Architecture.}
Planning and control run as independent processes and communicate through a publisher--subscriber protocol.
The planner replans at 2\,Hz from the latest hybrid history with the goal re-expressed in the current egocentric frame, and publishes the resulting reference motion for the controller to track at 50\,Hz; the realized robot state is then fed back to both the controller and planner.
During planner inference, the controller continues tracking the latest available reference and switches to the newly published trajectory once it arrives.
This fast-in-slow design enables non-blocking replanning while maintaining uninterrupted high-frequency control.

\paragraph{Latency Compensation via History Shift.}
To reduce the discontinuity caused by inference latency, RTC~\citep{black2026real,black2025training} notes that blending independently generated chunks can produce invalid intermediate actions and instead accounts for committed actions during inference or training. However, we regard it as a temporal misalignment of the autoregressive history and compensate for it through a predictive history shift. Let $d$ denote the expected number of motion steps before the new reference becomes available, estimated by an exponential moving average of recent inference times. We select the predicted history window $[t+d-H+1,t+d]$ and construct it bidirectionally from the latest realized state $\mathbf{s}_{t}^{\mathrm{real}}$, using the predicted relative kinematics to reconstruct the past and propagate forward toward the anticipated handoff state. The goal is then re-expressed relative to the anticipated handoff state $\mathbf{s}_{t+d}^{\mathrm{hyb}}$. This simple shift aligns both history and goal conditioning with the expected reference switch, reducing latency-induced handoff inconsistency without output blending, inference-time inpainting, or delay-aware retraining.

\section{Experiments}
\label{sec:experiments}
To systematically analyze our proposed method, we adopt a comprehensive experiments covering \textbf{Motion Quality and Executability}, \textbf{Goal-Reaching Precision}, and \textbf{Robustness and Recovery}, and answering the following questions:
(\textbf{Q1}) Can HIPC generate diverse and executable motions that can be tracked by a general controller?
(\textbf{Q2}) Can HIPC accurately reach different geometric goal conditions? 
(\textbf{Q3}) Is HIPC robust from out-of-distribution physical states?

\subsection{Experimental Setup and Real-World Deployment}

All experiments use the 29-DoF Unitree G1, with HIPC as the high-level planner and SONIC~\citep{luo2026sonic} as the downstream tracker.
HIPC replans asynchronously at 2\,Hz, while SONIC tracks the latest reference at 50\,Hz.
For comparison, each baseline uses the same SONIC tracker and 2\,Hz closed-loop protocol; evaluations are offline unless otherwise stated to avoid inference-latency effects.

For real-world deployment, HIPC and SONIC run off-board on an NVIDIA RTX 4070 GPU, with motion capture providing robot poses for egocentric goal construction.
Across five goal configurations with three trials each, HIPC responds online to goal switches and corrects spatial deviations, including target overshoot.
If the robot falls during locomotion, it can recover to standing and continue toward the goal without manual intervention.
Additional details are provided in the Appendix.


\subsection{Analysis for Motion Quality and Executability}
\label{sec:motion_quality}

To address \textbf{Q1}, we evaluate HIPC from two perspectives: the quality of the generated motions and their executed motions. The former focuses on replanning smoothness and geometric plausibility, while the latter evaluates whether the generated references remain stable, goal-directed, and semantically meaningful after physical execution. All metric definitions are provided in the Appendix.

\paragraph{Comparison on Motion Quality.}
We evaluate planning-motion quality on 100STYLE~\citep{mason2022local}, using sampled motion histories and future root positions as planar goals.
As shown in Table~\ref{tab:motion_quality}, HIPC substantially improves replanning smoothness and geometric quality, with lower boundary jerk, limited penetration, and reduced foot artifacts, while preserving motion diversity and text--motion alignment comparable to existing generative baselines.
These improvements also translate to physical execution: after SONIC tracking, HIPC achieves zero falls and the lowest endpoint error among goal-conditioned baselines ($0.101$\,m), together with the best FID.
Overall, the results indicate that HIPC produces references that are not only smoother across replanning boundaries, but also more physically executable without sacrificing the distributional and semantic properties of generated motion.

\begin{table*}[t]
\centering
\caption{
\textbf{Comparison of planning-motion quality and tracker-conditioned execution.}
``$\rightarrow$'' denotes closer agreement with the dataset reference.
\textbf{Bold} values denote the best results excluding the dataset.
Mean $\pm$ standard deviation over three rollout seeds.
}
\label{tab:motion_quality}

\setlength{\tabcolsep}{5pt}
\renewcommand{\arraystretch}{1.15}
\small

\textbf{(a) Replanning smoothness and geometric quality of planning motion}

\vspace{2pt}

\resizebox{\textwidth}{!}{
\begin{tabular}{@{}lcccccccc@{}}
\toprule
\textbf{Method}
&
\multicolumn{2}{c}{\textbf{Replanning Smoothness}}
&
\multicolumn{2}{c}{\textbf{Ground Penetration}}
&
\multicolumn{2}{c}{\textbf{Self-Penetration}}
&
\multicolumn{2}{c}{\textbf{Foot--Ground Contact}}
\\
\cmidrule(lr){2-3}
\cmidrule(lr){4-5}
\cmidrule(lr){6-7}
\cmidrule(lr){8-9}
&
\textbf{PJ} $\rightarrow$
&
\textbf{AUJ} $\downarrow$
&
\textbf{Time (\%)} $\downarrow$
&
\textbf{Depth (cm)} $\downarrow$
&
\textbf{Time (\%)} $\downarrow$
&
\textbf{Depth (cm)} $\downarrow$
&
\textbf{Skate (cm/s)} $\downarrow$
&
\textbf{Float (cm)} $\downarrow$
\\
\midrule

Dataset
& $0.166\std{0.717}$
& $0.915\std{1.420}$
& $93.000\std{14.551}$
& $2.963\std{0.784}$
& $29.440\std{37.016}$
& $3.119\std{1.688}$
& $0.886\std{1.631}$
& $0.139\std{1.058}$
\\

\midrule

Kimodo
& $0.982\std{0.778}$
& $4.991\std{3.505}$
& $44.427\std{19.989}$
& $2.462\std{1.697}$
& $9.710\std{14.081}$
& $2.981\std{1.604}$
& $\mathbf{6.885\std{76.076}}$
& $0.129\std{1.368}$
\\

ARDY
& $0.822\std{0.482}$
& $4.170\std{2.883}$
& $\mathbf{28.057\std{11.985}}$
& $2.291\std{1.217}$
& $15.250\std{17.508}$
& $\mathbf{2.964\std{1.535}}$
& $29.663\std{117.040}$
& $0.445\std{2.240}$
\\

TextOp
& $1.003\std{0.959}$
& $6.946\std{6.606}$
& $70.987\std{21.212}$
& $5.163\std{5.111}$
& $34.243\std{27.974}$
& $3.831\std{2.146}$
& $103.696\std{288.884}$
& $1.146\std{4.905}$
\\

OMG
& $0.915\std{0.461}$
& $6.388\std{2.847}$
& $52.363\std{19.597}$
& $2.909\std{3.089}$
& $30.160\std{27.405}$
& $3.685\std{1.884}$
& $85.147\std{163.188}$
& $0.940\std{3.133}$
\\

\textbf{HIPC} (Ours)
& $\mathbf{0.492\std{0.264}}$
& $\mathbf{2.894\std{1.166}}$
& $90.920\std{13.026}$
& $\mathbf{2.080\std{0.730}}$
& $\mathbf{5.350\std{17.259}}$
& $3.773\std{2.104}$
& $17.378\std{62.129}$
& $\mathbf{0.042\std{0.380}}$
\\

\bottomrule
\end{tabular}
}

\vspace{8pt}

\textbf{(b) Goal-reaching, distributional, and semantic quality of tracking result}

\vspace{2pt}

\resizebox{\textwidth}{!}{
\begin{tabular}{@{}lcccccccc@{}}
\toprule
\textbf{Method}
&
\multicolumn{2}{c}{\textbf{Execution}}
&
\multicolumn{2}{c}{\textbf{Motion Distribution}}
&
\multicolumn{4}{c}{\textbf{Text--Motion Alignment}}
\\
\cmidrule(lr){2-3}
\cmidrule(lr){4-5}
\cmidrule(lr){6-9}
&
\textbf{Fall (\%)} $\downarrow$
&
\textbf{$E_{\mathrm{pos}}$ (m)} $\downarrow$
&
\textbf{FID} $\downarrow$
&
\textbf{DIV} $\rightarrow$
&
\textbf{R@1} $\uparrow$
&
\textbf{R@2} $\uparrow$
&
\textbf{R@3} $\uparrow$
&
\textbf{MM-Dist} $\downarrow$
\\
\midrule

Dataset
& --
& --
& $0.000\std{0.000}$
& $1.188\std{0.008}$
& $0.178\std{0.022}$
& $0.270\std{0.021}$
& $0.339\std{0.017}$
& $1.301\std{0.080}$
\\

\midrule

Kimodo
& $7.000\std{25.515}$
& $0.457\std{0.408}$
& $0.621\std{0.001}$
& $\mathbf{1.190\std{0.007}}$
& $\mathbf{0.132\std{0.024}}$
& $\mathbf{0.210\std{0.024}}$
& $\mathbf{0.286\std{0.023}}$
& $1.329\std{0.002}$
\\

ARDY
& $1.667\std{12.802}$
& $0.242\std{0.238}$
& $0.605\std{0.006}$
& $1.228\std{0.005}$
& $0.100\std{0.012}$
& $0.173\std{0.024}$
& $0.241\std{0.024}$
& $1.338\std{0.002}$
\\

TextOp
& $24.667\std{43.107}$
& --
& $0.743\std{0.009}$
& $1.238\std{0.002}$
& $0.079\std{0.005}$
& $0.157\std{0.000}$
& $0.231\std{0.006}$
& $1.333\std{0.000}$
\\

OMG
& $7.000\std{25.515}$
& --
& $0.747\std{0.032}$
& $1.194\std{0.012}$
& $0.075\std{0.011}$
& $0.139\std{0.016}$
& $0.203\std{0.022}$
& $\mathbf{1.322\std{0.012}}$
\\

\textbf{HIPC} (Ours)
& $\mathbf{0.000\std{0.000}}$
& $\mathbf{0.101\std{0.067}}$
& $\mathbf{0.574\std{0.000}}$
& $1.194\std{0.000}$
& $0.106\std{0.000}$
& $0.175\std{0.000}$
& $0.236\std{0.000}$
& $1.324\std{0.000}$
\\

\bottomrule
\end{tabular}
}

\end{table*}

\paragraph{Ablation on Motion Closure Mechanism.}
Table~\ref{tab:closed_compare} isolates the effects of execution feedback, history re-anchoring, and latency compensation.
Predicted history avoids execution-induced distribution shift but leaves tracking deviations unobserved, causing spatial drift, while directly using realized histories introduces dynamics-dependent mismatch and less smooth handoff.
History re-anchoring resolves this trade-off by anchoring planner-native history to the realized state while preserving its kinematic structure, substantially improving goal accuracy.
The additional reconstruction increases average end-to-end planning latency from 92\,ms to 120\,ms, still within the 2\,Hz replanning budget.
Latency compensation then accounts for state evolution during this interval, recovering smooth and stable handoff.
Together, these results validate execution-consistent history re-anchoring and latency-aware replanning for closing the planning--control loop.
\begin{table*}[t]
\centering
\caption{
\textbf{Ablation of motion-closure mechanisms.}
We compare planner-history constructions and handoff strategies that isolate execution feedback, history re-anchoring, and latency compensation.
Metrics evaluate goal accuracy, stability, and seam smoothness; results are mean$\pm$std over three seeds.
}
\label{tab:closed_compare}

\setlength{\tabcolsep}{6pt}
\renewcommand{\arraystretch}{1.15}
\scriptsize

\begin{tabular}{@{}llccc@{}}
\toprule
\multicolumn{2}{c}{\textbf{Iterative Design}}
&
\multicolumn{2}{c}{\textbf{Goal Reaching}}
&
\textbf{Smoothness}
\\
\cmidrule(lr){1-2}
\cmidrule(lr){3-4}
\cmidrule(lr){5-5}

\textbf{Planner History}
&
\textbf{Chunk Handoff}
&
\textbf{$E_{\mathrm{pos}}$ (m)} $\downarrow$
&
\textbf{Fall (\%)} $\downarrow$
&
\textbf{Seam PJ} $\downarrow$
\\
\midrule

Predicted
& Direct Switching
& $0.371\std{0.037}$
& $0.000\std{0.000}$
& $0.005\std{0.000}$
\\

Realized
& Direct Switching
& $0.547\std{0.093}$
& $4.444\std{3.143}$
& $0.012\std{0.002}$
\\


Hybrid
& Direct Switching
& $0.147\std{0.028}$
& $2.222\std{3.143}$
& $0.009\std{0.002}$
\\


\textbf{Hybrid}
& \textbf{Latency Compensation}
& $\mathbf{0.129}\std{0.017}$
& $\mathbf{0.000}\std{0.000}$
& $\mathbf{0.005}\std{0.000}$
\\

\bottomrule
\end{tabular}
\vspace{-5pt}
\end{table*}


\subsection{Analysis for Goal Reaching Ability}

To answer \textbf{Q2}, we evaluate HIPC from two complementary aspects: unified realization of diverse static goal conditions and online tracking of dynamically updated goals from upstream planners.

\paragraph{Static Goal Achievement.}
We evaluate HIPC on 330 trials covering 11 goal configurations, with 30 trials per setting and states and targets sampled from 100STYLE~\citep{mason2022local}.
Table~\ref{tab:goal_reaching} shows that a single planner supports root position, orientation, velocity, joint, and end-effector goals, together with representative compositions.
While individual goals are matched accurately, composed conditions exhibit cross-goal interference: increasing the number of constraints does not consistently improve each component and can reduce its accuracy as multiple objectives compete within the same motion.
This trade-off is most apparent for full-state goals, and adding text introduces a further modest degradation, showing that semantic and geometric conditioning can be combined but are not completely decoupled.

\begin{figure}[b]
\vspace{-5pt}
    \centering
    \includegraphics[width=\textwidth]{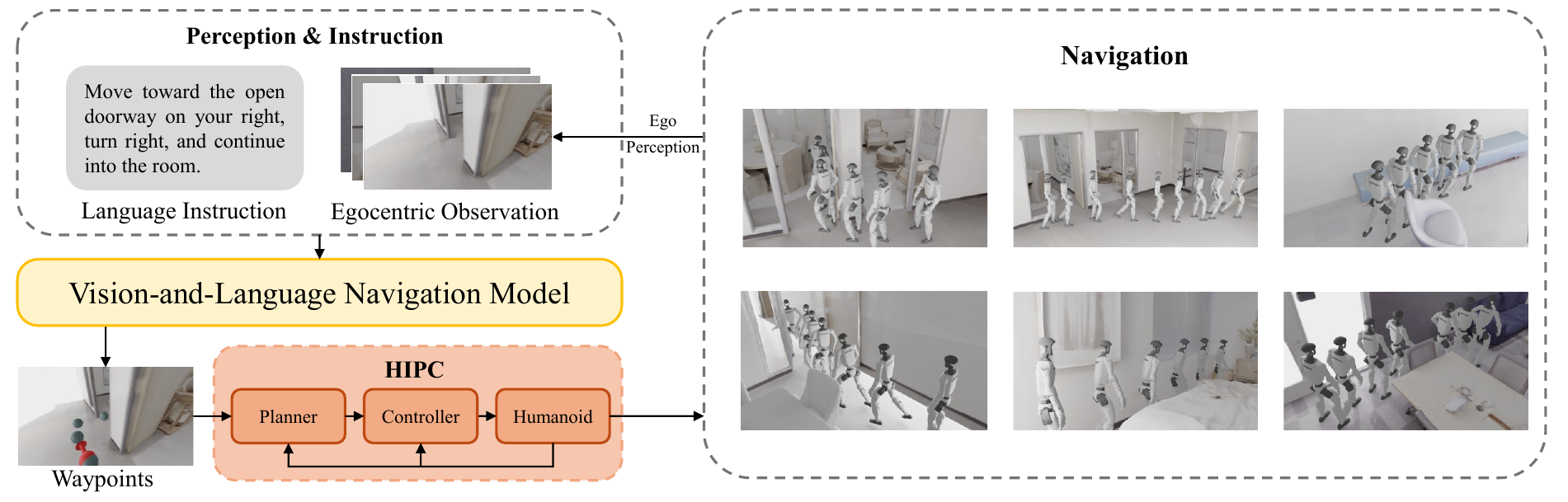}
    \caption{
        \textbf{Dynamic goal tracking for navigation.} HIPC executes dynamically updated waypoints from vision-and-language navigation planners as egocentric whole-body goals in Replica scenes.
    }
    \label{fig:vln}
\end{figure}

\paragraph{Dynamic Goal Tracking.}
As shown in Fig.~\ref{fig:vln}, we evaluate HIPC in six Replica indoor scenes~\citep{straub2019replica} imported into Isaac Sim, using two upstream navigation modules: an occupancy-map-based A* planner with obstacle inflation and the vision-and-language navigation model OmniNav~\citep{xue2025omninav}.
Their waypoints are transformed into HIPC's egocentric goal conditions for online navigation.
Across 50 navigation trials each, A* and OmniNav achieve 90\% and 70\% success in reaching the final navigation goal, respectively.
The gap reflects the greater reliability of map-based geometric planning, while the VLN result shows that HIPC can reliably track dynamically updated navigation goals from higher-level upstream planners, supporting its use as a planner-agnostic whole-body execution interface.

\begin{table*}[t]
\centering
\caption{
\textbf{Goal reaching under individual and representative composed conditions.}
For individual goals, ``Error'' denotes the corresponding goal-specific metric.
Representative compositions are evaluated jointly over all specified components;
additional combinations are reported in the Appendix.
}
\label{tab:goal_reaching}

\setlength{\tabcolsep}{4pt}
\renewcommand{\arraystretch}{1.12}
\scriptsize

\begin{minipage}[t]{0.30\textwidth}
\centering
\textbf{(a) Individual goals}

\vspace{2pt}

\begin{tabular}{@{}lc@{}}
\toprule
\textbf{Goal Condition}
&
\textbf{Error}
\\
\midrule
Root Position
& $0.079\std{0.049}$
\\
Root Orientation
& $0.092\std{0.026}$
\\
Joint Configuration
& $0.129\std{0.085}$
\\
Root Velocity
& $0.361\std{0.303}$
\\
End-Effector Position
& $0.156\std{0.051}$
\\
\bottomrule
\end{tabular}
\end{minipage}
\hfill
\begin{minipage}[t]{0.68\textwidth}
\centering
\textbf{(b) Composed goals}

\vspace{2pt}

\resizebox{\linewidth}{!}{
\begin{tabular}{@{}lccccc@{}}
\toprule
\textbf{Goal Condition}
&
\textbf{$\Epos$ (m)} $\downarrow$
&
\textbf{$\Eori$ (rad)} $\downarrow$
&
\textbf{$E_{\mathrm{MPJPE}}^{\mathrm{goal}}$ (m)} $\downarrow$
&
\textbf{$\Evel$ (m/s)} $\downarrow$
&
\textbf{$E_{\mathrm{ee}}$ (m)} $\downarrow$
\\
\midrule
Position + Orientation
& $0.110\std{0.069}$ & $\mathbf{0.082}\std{0.044}$ & -- & -- & --
\\
Position + Joint
& $0.112\std{0.060}$ & -- & $0.116\std{0.081}$ & -- & --
\\
Position + Velocity
& $0.095\std{0.061}$ & -- & -- & $\mathbf{0.576}\std{0.469}$ & --
\\
Position + End-Effector
& $\mathbf{0.085}\std{0.038}$ & -- & -- & -- & $\mathbf{0.146}\std{0.035}$
\\
Full-state Goal
& $0.130\std{0.063}$ & $0.133\std{0.053}$ & $\mathbf{0.056}\std{0.016}$ & $0.577\std{0.470}$ & $0.182\std{0.076}$
\\
Full-state Goal + Text
& $0.142\std{0.137}$ & $0.166\std{0.121}$ & $0.069\std{0.029}$ & $0.579\std{0.455}$ & $0.206\std{0.142}$
\\
\bottomrule
\end{tabular}
}
\end{minipage}

\end{table*}

\subsection{Robustness and Recovery Analysis}
To answer \textbf{Q3}, we evaluate whether HIPC can close the planning--control loop under large execution deviations caused by transient perturbations and dynamically induced falls.
We assess its ability to preserve goal-directed motion after disturbances and recover from fallen states through endpoint error, fall rate, response smoothness, recovery success, and get-up time.

\paragraph{Robustness to Physical Perturbations.}
Following the same goal-reaching protocol as Sec.~\ref{sec:motion_quality}, we apply randomized external forces and evaluate post-perturbation stability, goal preservation, and response smoothness.
As shown in Table~\ref{tab:robustness_recovery}(a), HIPC achieves zero falls together with the lowest endpoint error and response PJ, demonstrating that execution-consistent replanning can correct disturbance-induced deviations while preserving both task accuracy and smooth physical execution.

\begin{table*}[t]
\centering
\caption{
\textbf{Robustness to physical perturbations and fall recovery.}
(a) evaluates goal reaching after randomized external disturbances.
(b) induces falls with a 500\,N horizontal force for 0.1\,s in a uniformly random direction, then commands a same-position standing goal and reports recovery rate and get-up time over successful trials.
}
\label{tab:robustness_recovery}

\setlength{\tabcolsep}{3.5pt}
\scriptsize

\begin{minipage}[t]{0.46\textwidth}
\vspace{0pt}
\centering
\textbf{(a) Robustness under perturbations}

\vspace{2pt}

\renewcommand{\arraystretch}{1.12}
\makebox[\linewidth][c]{%
\scalebox{0.88}{%
\begin{tabular}{@{}lccc@{}}
\toprule
\textbf{Method}
&
\textbf{Fall (\%)} $\downarrow$
&
\textbf{$E_{\mathrm{pos}}$ (m)} $\downarrow$
&
\textbf{Response PJ} $\downarrow$
\\
\midrule
TextOp
& $29.167\std{0.851}$
& $1.615\std{0.065}$
& $0.070\std{0.007}$
\\
Kimodo
& $7.986\std{2.140}$
& $0.443\std{0.014}$
& $0.059\std{0.005}$
\\
ARDY
& $2.431\std{1.299}$
& $0.251\std{0.022}$
& $0.045\std{0.005}$
\\
OMG
& $8.681\std{4.019}$
& $2.642\std{0.109}$
& $0.068\std{0.005}$
\\
\textbf{HIPC (Ours)}
& $\mathbf{0.000}\std{0.000}$
& $\mathbf{0.105}\std{0.005}$
& $\mathbf{0.034}\std{0.001}$
\\
\bottomrule
\end{tabular}%
}}
\end{minipage}
\hfill
%
\begin{minipage}[t]{0.52\textwidth}
\vspace{0pt}
\centering
\textbf{(b) Recovery after induced falls}

\vspace{2pt}

\renewcommand{\arraystretch}{1.38}
\makebox[\linewidth][c]{%
\scalebox{0.88}{%
\begin{tabular}{@{}lcc@{}}
\toprule
\textbf{Variant}
&
\textbf{Recovery (\%)} $\uparrow$
&
\textbf{$T_{\mathrm{getup}}$ (s)} $\downarrow$
\\
\midrule
Heading-based Representation
& $0.000\std{0.000}$
& --
\\
w/o History Re-anchoring
& $0.000\std{0.000}$
& --
\\
w/o Deviation-Decayed Recovery
& $0.000\std{0.000}$
& --
\\
\textbf{HIPC}
& $\mathbf{53.333}\std{9.428}$
& $\mathbf{7.227}\std{0.318}$
\\
\bottomrule
\end{tabular}%
}}
\end{minipage}
\vspace{-7pt}
\end{table*}

\paragraph{Fall Recovery.}
We induce diverse falls using a 500\,N horizontal force for 0.1\,s with random direction, then command an upright standing goal at the same position.
HIPC recovers in $53.33\%$ of trials with a mean get-up time of $7.23$\,s, whereas the heading-based representation and variants without history re-anchoring or deviation-decayed recovery achieve no successful recoveries (Table~\ref{tab:robustness_recovery}(b)).
This confirms the importance of yaw-free representation, execution-consistent history construction, and recovery-oriented training.

\section{Conclusion and Limitation}
\paragraph{Conclusion.}
We presented HIPC, a closed-loop humanoid planning-and-control framework that integrates goal-conditioned generative motion planning with execution-consistent feedback and latency-aware asynchronous control.
By closing the loop between motion generation and physical execution while preserving planner--controller modularity, HIPC supports accurate goal reaching, robust perturbation recovery, and recovery from unseen fallen configurations across diverse whole-body tasks.

\paragraph{Limitations.}
Despite these strengths, HIPC currently closes the loop primarily between motion planning and physical execution, while relying on externally provided goal states rather than direct environment perception. In real-world deployment, accurate spatial goals are therefore obtained from external localization systems such as motion capture, which increases sensing and communication complexity. 
Additionally, controller-dependent tracking errors may persist despite HIPC's closed-loop corrections.
Addressing these errors while preserving real-time performance and planner--controller modularity remains a direction for future work.


\newpage
\bibliography{iclr2027_conference}
\bibliographystyle{iclr2027_conference}

\newpage
\appendix
\section{Appendix}
\subsection{Motion Generation}
\paragraph{Horizontal Projection.}
The horizontal root increment is defined with respect to the instantaneous gravity direction rather than a yaw-aligned world plane. Given the local root displacement
\begin{equation}
{}^{E_{t-1}}\Delta\mathbf{p}_{t-1}
=
\mathbf{R}_{t-1}^{\top}
\left(
{}^W\mathbf{p}_{t}-{}^W\mathbf{p}_{t-1}
\right),
\end{equation}
we remove its component parallel to the local gravity vector
\(\mathbf{g}_{t-1}=\mathbf{R}_{t-1}^{\top}\mathbf{g}\).
The projected horizontal displacement is therefore
\begin{equation}
{}^{E_{t-1}}\Delta\mathbf{p}_{t-1}^{\mathrm h}
=
{}^{E_{t-1}}\Delta\mathbf{p}_{t-1}
-
\frac{
\left(
{}^{E_{t-1}}\Delta\mathbf{p}_{t-1}
\right)^\top
\mathbf{g}_{t-1}
}{
\|\mathbf{g}_{t-1}\|_2^2
}
\mathbf{g}_{t-1}.
\end{equation}
Equivalently, this applies the orthogonal projection matrix
\begin{equation}
\mathbf{P}_{t-1}^{\mathrm h}
=
\mathbf{I}
-
\frac{
\mathbf{g}_{t-1}\mathbf{g}_{t-1}^{\top}
}{
\|\mathbf{g}_{t-1}\|_2^2
}
\end{equation}
to the local displacement. This construction defines horizontal motion intrinsically from gravity and therefore remains valid even when the root is strongly tilted or inverted.

\paragraph{State--Motion Representation Conversion.}
Our canonical representation admits deterministic conversion between a state trajectory and its motion representation. Let a robot state be
\begin{equation}
\mathbf{s}_t=
\left[
{}^W\mathbf{p}_t,\,
\mathbf{R}_t,\,
\mathbf{q}_t,\,
\mathbf{b}_t
\right],
\end{equation}
and consider a sequence of \(T+1\) states
\(\mathbf{s}_{0:T}\). Applying Eq.~\ref{eq:motion_feature} to each consecutive pair
\((\mathbf{s}_{t-1},\mathbf{s}_t)\) produces \(T\) motion frames
\(\mathbf{x}_{1:T}\). Hence, in our planner, \(H+F+1\) states correspond to \(H+F\) motion frames.

The inverse conversion follows directly from the incremental construction. Given \(\mathbf{x}_{1:T}\) and the initial root pose
\(({}^W\mathbf{p}_0,\mathbf{R}_0)\), we recursively reconstruct the root orientation as
\begin{equation}
\mathbf{R}_t
=
\mathbf{R}_{t-1}\,
{}^{E_{t-1}}\mathbf{R}_{E_t}.
\end{equation}
The local horizontal displacement is transformed back to the world frame and accumulated, while the vertical coordinate is assigned directly from the predicted height:
\begin{equation}
\begin{aligned}
{}^W\Delta\mathbf{p}_{t-1}^{\mathrm h}
&=
\mathbf{R}_{t-1}\,
{}^{E_{t-1}}\Delta\mathbf{p}_{t-1}^{\mathrm h},\\
{}^W\mathbf{p}_t^{\mathrm h}
&=
{}^W\mathbf{p}_{t-1}^{\mathrm h}
+
{}^W\Delta\mathbf{p}_{t-1}^{\mathrm h},\\
({}^W\mathbf{p}_t)_z
&=
h_t.
\end{aligned}
\end{equation}
The joint configuration \(\mathbf{q}_t\) and contact state \(\mathbf{b}_t\) are stored explicitly in \(\mathbf{x}_t\), and \(\mathbf{g}_t\) is consistent with the reconstructed orientation through
\(\mathbf{g}_t=\mathbf{R}_t^\top\mathbf{g}\).
Consequently, \(\mathbf{x}_{1:T}\) together with the initial root pose uniquely determines the subsequent states \(\mathbf{s}_{1:T}\), up to numerical precision in the rotation representation.

The mapping is intentionally not bijective. When converting
\(\mathbf{s}_{0:T}\) to \(\mathbf{x}_{1:T}\), the initial joint and contact states
\((\mathbf{q}_0,\mathbf{b}_0)\) are not retained, and the global planar translation and heading are removed by the canonical representation. Therefore, recovering the original \(T+1\) states additionally requires the initial full state; with only the initial root pose, the remaining \(T\) states are reconstructible, whereas \(\mathbf{s}_0\) is not fully recoverable.

\paragraph{State Deviation and Composition.}
We define the operators $\ominus$ and $\oplus$ used in history re-anchoring directly in the robot state space.
For the root pose, let
\begin{equation}
\mathbf{T}(\mathbf{s})
=
\begin{bmatrix}
\mathbf{R} & {}^W\mathbf{p}\\
\mathbf{0}^{\top} & 1
\end{bmatrix}
\in \mathrm{SE}(3).
\end{equation}
Given two states $\mathbf{s}^a$ and $\mathbf{s}^b$, their state deviation
$\boldsymbol{\delta}=\mathbf{s}^a\ominus\mathbf{s}^b$ consists of the relative root transform and joint offset
\begin{equation}
\Delta\mathbf{T}
=
\mathbf{T}(\mathbf{s}^a)
\mathbf{T}(\mathbf{s}^b)^{-1},
\qquad
\Delta\mathbf{q}
=
\mathbf{q}^a-\mathbf{q}^b.
\end{equation}
Applying this deviation to another state $\mathbf{s}$ is defined by
\begin{equation}
\boldsymbol{\delta}\oplus\mathbf{s}
=
\left[
\Delta\mathbf{T}\,\mathbf{T}(\mathbf{s}),
\;
\mathbf{q}+\Delta\mathbf{q},
\;
\mathbf{b}
\right],
\end{equation}
where the first term denotes rigid-body composition of the root pose and the contact state is inherited from $\mathbf{s}$.
By construction,
\begin{equation}
\left(
\mathbf{s}^a\ominus\mathbf{s}^b
\right)
\oplus
\mathbf{s}^b
=
\mathbf{s}^a,
\end{equation}
for the root pose and joint configuration.
In Eq.~\ref{eq:history_reanchoring}, the deviation measured at the current frame is applied uniformly to every planner-generated state in the history, yielding a coherent rigid-body displacement and joint-space offset rather than independently replacing individual historical states.

\paragraph{Losses.}
The VAE is trained with feature reconstruction, latent regularization, and kinematic supervision:
\begin{equation}
\mathcal{L}_{\mathrm{VAE}}
=
\lambda_{\mathrm{rec}}\mathcal{L}_{\mathrm{rec}}
+
\lambda_{\mathrm{KL}}\mathcal{L}_{\mathrm{KL}}
+
\mathcal{L}_{\mathrm{geo}},
\end{equation}
where $\mathcal{L}_{\mathrm{rec}}$ is a Huber loss on the canonical motion features and $\mathcal{L}_{\mathrm{KL}}$ regularizes the latent posterior toward a standard Gaussian. The geometric loss is
\begin{equation}
\begin{aligned}
\mathcal{L}_{\mathrm{geo}}
={}&
\lambda_{\mathrm{trans}}\mathcal{L}_{\mathrm{body}}^{\mathrm{trans}}
+
\lambda_{\mathrm{rot}}\mathcal{L}_{\mathrm{body}}^{\mathrm{rot}}
+
\lambda_{\mathrm{dof}}\mathcal{L}_{\mathrm{dof}}
+
\lambda_{\mathrm{vel}}\mathcal{L}_{\mathrm{dof}}^{\mathrm{vel}}\\
&+
\lambda_{\mathrm{contact}}\mathcal{L}_{\mathrm{contact}}
+
\lambda_{\mathrm{chord}}\mathcal{L}_{\mathrm{chord}} .
\end{aligned}
\end{equation}
The first four terms supervise FK-space body poses, joint configurations, and angular velocities, while $\mathcal{L}_{\mathrm{contact}}$ penalizes support-foot velocity during contact. Euclidean quantities use the Huber loss, and body orientations use a chordal rotation distance.

We further constrain the relative root rotation encoded by the representation directly on $\mathrm{SO}(3)$. After mapping the predicted 6D rotation to a rotation matrix,
\begin{equation}
\mathcal{L}_{\mathrm{chord}}
=
\frac{1}{F}
\sum_{t=1}^{F}
\left\|
\hat{\mathbf{R}}_{t}^{\mathrm{rel}}
-
\mathbf{R}_{t}^{\mathrm{rel}}
\right\|_F^2 .
\end{equation}
Unlike $\mathcal{L}_{\mathrm{body}}^{\mathrm{rot}}$, which supervises global orientations of all body links after forward kinematics, this term directly constrains the relative-root-rotation variable itself.

The diffusion planner is optimized with
\begin{equation}
\mathcal{L}_{\mathrm{LDM}}
=
\lambda_{\mathrm{simple}}\mathcal{L}_{\mathrm{simple}}
+
\lambda_{\mathrm{rec}}\mathcal{L}_{\mathrm{rec}}
+
\mathcal{L}_{\mathrm{geo}}
+
\mathcal{L}_{\mathrm{goal}},
\end{equation}
where $\mathcal{L}_{\mathrm{simple}}$ is the Huber denoising loss in latent space. For geometric conditioning, we reconstruct the state trajectory from the predicted canonical motion and evaluate the specified quantities at the goal frame $t^\star$, determined by the time-to-arrival condition when provided and otherwise set to the final predicted frame. The goal objective is
\begin{equation}
\begin{aligned}
\mathcal{L}_{\mathrm{goal}}
={}&
\lambda_{\mathrm{pos}}^{\mathrm{h}}\mathcal{L}_{\mathrm{pos}}^{\mathrm{h}}
+
\lambda_{\mathrm{pos}}^{\mathrm{v}}\mathcal{L}_{\mathrm{pos}}^{\mathrm{v}}
+
\lambda_{\mathrm{ori}}\mathcal{L}_{\mathrm{ori}}
+
\lambda_{\mathrm{g}}\mathcal{L}_{\mathrm{g}}\\
&+
\lambda_{\mathrm{joint}}\mathcal{L}_{\mathrm{joint}}
+
\lambda_{\mathrm{vel}}\mathcal{L}_{\mathrm{vel}}
+
\lambda_{\mathrm{ee}}\mathcal{L}_{\mathrm{ee}} .
\end{aligned}
\end{equation}
These terms supervise horizontal displacement, height, root orientation, gravity direction, joint configuration, root velocity, and four hand/foot end-effector positions, respectively. Each goal loss is applied only to samples for which the corresponding condition is enabled and retained after condition dropout, and its weight is selected from the locomotion or recovery loss profile.

\subsection{Motion Generation}
\paragraph{Horizontal Projection.}
The horizontal root increment is defined with respect to the instantaneous gravity direction rather than a yaw-aligned world plane. Given the local root displacement
\begin{equation}
{}^{E_{t-1}}\Delta\mathbf{p}_{t-1}
=
\mathbf{R}_{t-1}^{\top}
\left(
{}^W\mathbf{p}_{t}-{}^W\mathbf{p}_{t-1}
\right),
\end{equation}
we remove its component parallel to the local gravity vector
\(\mathbf{g}_{t-1}=\mathbf{R}_{t-1}^{\top}\mathbf{g}\).
The projected horizontal displacement is therefore
\begin{equation}
{}^{E_{t-1}}\Delta\mathbf{p}_{t-1}^{\mathrm h}
=
{}^{E_{t-1}}\Delta\mathbf{p}_{t-1}
-
\frac{
\left(
{}^{E_{t-1}}\Delta\mathbf{p}_{t-1}
\right)^\top
\mathbf{g}_{t-1}
}{
\|\mathbf{g}_{t-1}\|_2^2
}
\mathbf{g}_{t-1}.
\end{equation}
Equivalently, this applies the orthogonal projection matrix
\begin{equation}
\mathbf{P}_{t-1}^{\mathrm h}
=
\mathbf{I}
-
\frac{
\mathbf{g}_{t-1}\mathbf{g}_{t-1}^{\top}
}{
\|\mathbf{g}_{t-1}\|_2^2
}
\end{equation}
to the local displacement. This construction defines horizontal motion intrinsically from gravity and therefore remains valid even when the root is strongly tilted or inverted.

\paragraph{State--Motion Representation Conversion.}
Our canonical representation admits deterministic conversion between a state trajectory and its motion representation. Let a robot state be
\begin{equation}
\mathbf{s}_t=
\left[
{}^W\mathbf{p}_t,\,
\mathbf{R}_t,\,
\mathbf{q}_t,\,
\mathbf{b}_t
\right],
\end{equation}
and consider a sequence of \(T+1\) states
\(\mathbf{s}_{0:T}\). Applying Eq.~\ref{eq:motion_feature} to each consecutive pair
\((\mathbf{s}_{t-1},\mathbf{s}_t)\) produces \(T\) motion frames
\(\mathbf{x}_{1:T}\). Hence, in our planner, \(H+F+1\) states correspond to \(H+F\) motion frames.

The inverse conversion follows directly from the incremental construction. Given \(\mathbf{x}_{1:T}\) and the initial root pose
\(({}^W\mathbf{p}_0,\mathbf{R}_0)\), we recursively reconstruct the root orientation as
\begin{equation}
\mathbf{R}_t
=
\mathbf{R}_{t-1}\,
{}^{E_{t-1}}\mathbf{R}_{E_t}.
\end{equation}
The local horizontal displacement is transformed back to the world frame and accumulated, while the vertical coordinate is assigned directly from the predicted height:
\begin{equation}
\begin{aligned}
{}^W\Delta\mathbf{p}_{t-1}^{\mathrm h}
&=
\mathbf{R}_{t-1}\,
{}^{E_{t-1}}\Delta\mathbf{p}_{t-1}^{\mathrm h},\\
{}^W\mathbf{p}_t^{\mathrm h}
&=
{}^W\mathbf{p}_{t-1}^{\mathrm h}
+
{}^W\Delta\mathbf{p}_{t-1}^{\mathrm h},\\
({}^W\mathbf{p}_t)_z
&=
h_t.
\end{aligned}
\end{equation}
The joint configuration \(\mathbf{q}_t\) and contact state \(\mathbf{b}_t\) are stored explicitly in \(\mathbf{x}_t\), and \(\mathbf{g}_t\) is consistent with the reconstructed orientation through
\(\mathbf{g}_t=\mathbf{R}_t^\top\mathbf{g}\).
Consequently, \(\mathbf{x}_{1:T}\) together with the initial root pose uniquely determines the subsequent states \(\mathbf{s}_{1:T}\), up to numerical precision in the rotation representation.

The mapping is intentionally not bijective. When converting
\(\mathbf{s}_{0:T}\) to \(\mathbf{x}_{1:T}\), the initial joint and contact states
\((\mathbf{q}_0,\mathbf{b}_0)\) are not retained, and the global planar translation and heading are removed by the canonical representation. Therefore, recovering the original \(T+1\) states additionally requires the initial full state; with only the initial root pose, the remaining \(T\) states are reconstructible, whereas \(\mathbf{s}_0\) is not fully recoverable.

\paragraph{Losses.}
The VAE is trained with feature reconstruction, latent regularization, and kinematic supervision:
\begin{equation}
\mathcal{L}_{\mathrm{VAE}}
=
\lambda_{\mathrm{rec}}\mathcal{L}_{\mathrm{rec}}
+
\lambda_{\mathrm{KL}}\mathcal{L}_{\mathrm{KL}}
+
\mathcal{L}_{\mathrm{geo}},
\end{equation}
where $\mathcal{L}_{\mathrm{rec}}$ is a Huber loss on the canonical motion features and $\mathcal{L}_{\mathrm{KL}}$ regularizes the latent posterior toward a standard Gaussian. The geometric loss is
\begin{equation}
\begin{aligned}
\mathcal{L}_{\mathrm{geo}}
={}&
\lambda_{\mathrm{trans}}\mathcal{L}_{\mathrm{body}}^{\mathrm{trans}}
+
\lambda_{\mathrm{rot}}\mathcal{L}_{\mathrm{body}}^{\mathrm{rot}}
+
\lambda_{\mathrm{dof}}\mathcal{L}_{\mathrm{dof}}
+
\lambda_{\mathrm{vel}}\mathcal{L}_{\mathrm{dof}}^{\mathrm{vel}}\\
&+
\lambda_{\mathrm{contact}}\mathcal{L}_{\mathrm{contact}}
+
\lambda_{\mathrm{chord}}\mathcal{L}_{\mathrm{chord}} .
\end{aligned}
\end{equation}
The first four terms supervise FK-space body poses, joint configurations, and angular velocities, while $\mathcal{L}_{\mathrm{contact}}$ penalizes support-foot velocity during contact. Euclidean quantities use the Huber loss, and body orientations use a chordal rotation distance.

We further constrain the relative root rotation encoded by the representation directly on $\mathrm{SO}(3)$. After mapping the predicted 6D rotation to a rotation matrix,
\begin{equation}
\mathcal{L}_{\mathrm{chord}}
=
\frac{1}{F}
\sum_{t=1}^{F}
\left\|
\hat{\mathbf{R}}_{t}^{\mathrm{rel}}
-
\mathbf{R}_{t}^{\mathrm{rel}}
\right\|_F^2 .
\end{equation}
Unlike $\mathcal{L}_{\mathrm{body}}^{\mathrm{rot}}$, which supervises global orientations of all body links after forward kinematics, this term directly constrains the relative-root-rotation variable itself.

The diffusion planner is optimized with
\begin{equation}
\mathcal{L}_{\mathrm{LDM}}
=
\lambda_{\mathrm{simple}}\mathcal{L}_{\mathrm{simple}}
+
\lambda_{\mathrm{rec}}\mathcal{L}_{\mathrm{rec}}
+
\mathcal{L}_{\mathrm{geo}}
+
\mathcal{L}_{\mathrm{goal}},
\end{equation}
where $\mathcal{L}_{\mathrm{simple}}$ is the Huber denoising loss in latent space. For geometric conditioning, we reconstruct the state trajectory from the predicted canonical motion and evaluate the specified quantities at the goal frame $t^\star$, determined by the time-to-arrival condition when provided and otherwise set to the final predicted frame. The goal objective is
\begin{equation}
\begin{aligned}
\mathcal{L}_{\mathrm{goal}}
={}&
\lambda_{\mathrm{pos}}^{\mathrm{h}}\mathcal{L}_{\mathrm{pos}}^{\mathrm{h}}
+
\lambda_{\mathrm{pos}}^{\mathrm{v}}\mathcal{L}_{\mathrm{pos}}^{\mathrm{v}}
+
\lambda_{\mathrm{ori}}\mathcal{L}_{\mathrm{ori}}
+
\lambda_{\mathrm{g}}\mathcal{L}_{\mathrm{g}}\\
&+
\lambda_{\mathrm{joint}}\mathcal{L}_{\mathrm{joint}}
+
\lambda_{\mathrm{vel}}\mathcal{L}_{\mathrm{vel}}
+
\lambda_{\mathrm{ee}}\mathcal{L}_{\mathrm{ee}} .
\end{aligned}
\end{equation}
These terms supervise horizontal displacement, height, root orientation, gravity direction, joint configuration, root velocity, and four hand/foot end-effector positions, respectively. Each goal loss is applied only to samples for which the corresponding condition is enabled and retained after condition dropout, and its weight is selected from the locomotion or recovery loss profile.

\subsection{Evaluation Metrics}

\paragraph{Motion Quality.}
Motion quality is assessed in the feature space of a G1-specific
text--motion contrastive evaluator inspired by TMR.
The evaluator maps a two-second motion segment into a $512$-dimensional
$\ell_2$-normalized embedding: the $30$ G1 body positions are expressed in the
local root frame, sampled at $30\,\mathrm{Hz}$ over $60$ frames, and the root
translation and heading of the first frame are canonicalized away.
Text is encoded by a frozen T5-3B encoder followed by a learned projection into
the same feature space.

Every cohort is embedded by this single fixed evaluator, so the three cohorts
of a case are directly comparable: the \emph{ground truth} is the future source
window that follows the conditioning history, whereas the \emph{reference} and
\emph{executed} cohorts are the first two seconds of the planned and of the
physically executed trajectory of the same case. Source windows are recorded at
$50\,\mathrm{Hz}$ and resampled to the evaluator's $30\,\mathrm{Hz}$ input
contract. Since the learned feature space is specific to this evaluator,
absolute values are comparable across methods within a table but not with
numbers reported under a different evaluator.
Unless stated otherwise, the metrics below are evaluated over the 2-second
window that follows the conditioning history and does not contain the history
frames themselves.

\begin{itemize}
    \item \textbf{Frechet Inception Distance (FID).}
    FID is the principal metric for assessing the distributional similarity
    between generated and real motions in the extracted feature space.
    Let $(\mu_g, \Sigma_g)$ and $(\mu_x, \Sigma_x)$ be the means and
    covariances of the ground-truth and of a candidate cohort of embeddings.
    The FID is computed as
    \begin{equation}
        \mathrm{FID}
        =
        \|\mu_g-\mu_x\|_2^2
        +
        \operatorname{Tr}
        \left(
            \Sigma_g+\Sigma_x
            -2(\Sigma_g\Sigma_x)^{1/2}
        \right),
        \label{eq:fid}
    \end{equation}
    and is reported as $\mathrm{FID}(G,R)$ for the planning reference and
    $\mathrm{FID}(G,E)$ for the executed motion.
    It is one dataset-level statistic rather than a mean of per-case values,
    and $\mathrm{FID}(G,G)=0$ is reported as the dataset sanity row.

    \item \textbf{Diversity (DIV).}
    Diversity measures the spread of a cohort in the feature space.
    Randomly sampled pairs of distinct motion embeddings are compared, and
    their Euclidean distances are averaged:
    \begin{equation}
        \mathrm{DIV}
        =
        \frac{1}{P}
        \sum_{p=1}^{P}
        \left\|
            x_{a_p}-x_{b_p}
        \right\|_2
        ,\qquad a_p\neq b_p ,
        \label{eq:diversity}
    \end{equation}
    with $P=300$ sampled pairs.
    The quantity to match is the diversity of the dataset, so DIV is reported
    for the dataset, the reference, and the executed cohort and is scored by
    its agreement with the dataset value; insufficient and excessive diversity
    are both errors.

    \item \textbf{R-precision (R@K).}
    R@K evaluates text-to-motion semantic alignment by retrieval.
    A candidate batch of $M=32$ motion--text pairs is formed, each text is used
    as a query, and the $M$ motion embeddings of the batch are ranked by
    Euclidean distance to the query embedding.
    A query is a hit at $K$ when its paired motion is among the $K$ nearest
    candidates, and
    \begin{equation}
        \mathrm{R@}K
        =
        \frac{1}{M}
        \sum_{i=1}^{M}
        \mathbb{I}
        \left(
            \mathrm{rank}_i\le K
        \right)
        \label{eq:r_precision}
    \end{equation}
    is the mean hit rate over the batch.
    Because retrieval is defined text-to-motion, each text is the query and
    each candidate is a motion; the transposed motion-to-text score is a
    different quantity and is not reported.
    The final incomplete batch is completed with real negative motions so that
    every scored query is ranked against $32$ real candidates;
    zero-padded candidates are invalid because real embeddings are unit
    normalized, which would place every dummy at a fixed distance and alter the
    ranking.
    The reported value averages $20$ deterministic batch permutations.

    \item \textbf{Multimodal Distance (MM-Dist).}
    MM-Dist directly measures the average proximity between a motion and its
    corresponding text in the feature space.
    For $N$ valid text--motion pairs, let $\{h_i\}_{i=1}^{N}$ and
    $\{x_i\}_{i=1}^{N}$ be their respective text and motion feature vectors.
    MM-Dist is computed as
    \begin{equation}
        \mathrm{MM\text{-}Dist}
        =
        \frac{1}{N}
        \sum_{i=1}^{N}
        \|x_i-h_i\|_2.
        \label{eq:mm_dist}
    \end{equation}
    Despite its name, it is a paired text--motion distance rather than a
    distributional or a conditional-diversity statistic.

    \item \textbf{Multimodality (MMod.).}
    Multimodality measures the diversity of motions generated under the same
    text condition.
    For each of $C$ conditions, all distinct pairs of embeddings produced by
    independent repetitions of that condition are compared and averaged, and
    the condition-level values are then averaged:
    \begin{equation}
        \mathrm{MMod.}
        =
        \frac{1}{C}
        \sum_{c=1}^{C}
        \frac{1}{P_c}
        \sum_{p=1}^{P_c}
        \left\|
            x_{c,a_p}-x_{c,b_p}
        \right\|_2 .
        \label{eq:multimodality}
    \end{equation}
    MMod is well defined only across independently generated samples.
    Every run reads the same source segment, so the apparent cross-run
    ground-truth samples are duplicate copies of one motion whose pairwise
    distance is mechanically zero; the dataset MMod is therefore not reported.
    Since multimodality reflects conditional motion diversity rather than a
    quantity to be maximized unconditionally, the reference and executed
    cohorts are evaluated by their agreement with the dataset reference.

    \item \textbf{Peak Jerk (PJ).}
    PJ captures the most extreme instantaneous jerk occurring in a motion.
    Let $\mathbf{p}_t^{b}\in\mathbb{R}^3$ be the world position of body $b$ at
    frame $t$, obtained by forward kinematics from the evaluated joint
    configuration, and let
    \begin{equation}
        \mathbf{j}_t^{b}
        =
        \Delta^{3}\mathbf{p}_t^{b}
        =
        \mathbf{p}_{t}^{b}-3\mathbf{p}_{t-1}^{b}
        +3\mathbf{p}_{t-2}^{b}-\mathbf{p}_{t-3}^{b}
    \end{equation}
    be its third finite difference.
    Following the FlowMDM convention, the jerk is not divided by
    $\Delta t^{3}$, so the metric is expressed in
    $\mathrm{m}/\mathrm{frame}^{3}$.
    The framewise jerk magnitude is the largest $\ell_1$ jerk over all $B=30$
    bodies, and PJ is the mean over the $N$ evaluated cases of the largest
    framewise magnitude within the evaluation window $\mathcal{T}_i$:
    \begin{equation}
        \mathrm{PJ}
        =
        \frac{1}{N}
        \sum_{i=1}^{N}
        \max_{t\in\mathcal{T}_i}
        \max_{1\le b\le B}
        \left\|
            \mathbf{j}_{t,i}^{b}
        \right\|_1 .
        \label{eq:pj}
    \end{equation}

    \item \textbf{Area Under the Jerk (AUJ).}
    AUJ measures the cumulative deviation of a motion's smoothness from the
    dataset level.
    Let $j_{\mathrm{avg}}$ be the mean ground-truth framewise jerk magnitude,
    that is, the average of $\max_{b}\|\mathbf{j}_t^{b}\|_1$ over all
    ground-truth evaluation windows, and let
    $J_t=\max_{b}\|\mathbf{j}_t^{b}\|_1$ denote the framewise jerk magnitude of
    the evaluated motion.
    The metric is computed as
    \begin{equation}
        \mathrm{AUJ}
        =
        \sum_{t\in\mathcal{T}}
        \left|
            J_t-j_{\mathrm{avg}}
        \right| .
        \label{eq:auj}
    \end{equation}
    AUJ is a per-case quantity that is averaged over cases;
    exceeding the dataset smoothness and being smoother than it are both
    penalized.

    \item \textbf{Seam PJ and Response PJ.}
    The same jerk profile is also evaluated over short windows that isolate a
    discontinuity.
    \emph{Seam PJ} centers a window of $\pm10$ frames around every replanning
    seam, takes the largest framewise jerk magnitude within that window, and
    averages over the seams of the episode; the initial plan, which starts from
    a hold, is not a seam.
    \emph{Response PJ} takes the largest framewise jerk magnitude over the
    fixed 2-second window that begins at the onset of the external
    perturbation, evaluated on the physically executed motion.

    \item \textbf{Ground Penetration.}
    Ground penetration measures geometric violations between the generated
    robot motion and the ground plane.
    The floor is the plane geom of the robot model, whose surface lies at
    $z=0$.
    For every non-visual collision geom $q$ of the model, let $s_{t,q}$ be its
    signed distance to the floor at frame $t$;
    the maximum penetration depth at frame $t$ is
    \begin{equation}
        d_t^{\mathrm{GP}}
        =
        \max_{q}
        \left[
            -s_{t,q}
        \right]_+,
        \qquad
        [x]_+=\max(x,0).
        \label{eq:ground_penetration_frame}
    \end{equation}
    A frame counts as penetrating only when $d_t^{\mathrm{GP}}$ strictly
    exceeds $\tau_p=1\,\mathrm{cm}$, which suppresses contact noise.
    The penetration time is the percentage of penetrating frames,
    \begin{equation}
        \mathrm{GP}_{\mathrm{time}}
        =
        \frac{100}{T}
        \sum_{t=1}^{T}
        \mathbb{I}
        \left(
            d_t^{\mathrm{GP}}>\tau_p
        \right),
        \label{eq:ground_penetration_time}
    \end{equation}
    and the penetration depth is averaged over the penetrating frames only,
    \begin{equation}
        \mathrm{GP}_{\mathrm{depth}}
        =
        \frac{
            \sum_{t=1}^{T}
            d_t^{\mathrm{GP}}
            \mathbb{I}(d_t^{\mathrm{GP}}>\tau_p)
        }{
            \sum_{t=1}^{T}
            \mathbb{I}(d_t^{\mathrm{GP}}>\tau_p)
        },
        \label{eq:ground_penetration_depth}
    \end{equation}
    reported in centimeters.
    A penetration depth of $0\pm0\,\mathrm{cm}$ is reported when no frame
    violates the threshold.

    \item \textbf{Self-Penetration.}
    Self-penetration measures intersections between robot bodies that are not
    kinematically adjacent.
    Let $\mathcal{P}$ be the set of collision-geom pairs whose bodies belong to
    different branches of the kinematic tree, excluding every pair in which one
    body is an ancestor of the other, and let $s_{t,(i,j)}$ be their signed
    distance at frame $t$.
    The maximum self-penetration depth is
    \begin{equation}
        d_t^{\mathrm{SP}}
        =
        \max_{(i,j)\in\mathcal{P}}
        \left[
            -s_{t,(i,j)}
        \right]_+ .
        \label{eq:self_penetration_frame}
    \end{equation}
    The self-penetration time and depth apply the same
    $\tau_p=1\,\mathrm{cm}$ violation threshold and the same averaging rule as
    ground penetration,
    \begin{equation}
        \mathrm{SP}_{\mathrm{time}}
        =
        \frac{100}{T}
        \sum_{t=1}^{T}
        \mathbb{I}
        \left(
            d_t^{\mathrm{SP}}>\tau_p
        \right),
        \qquad
        \mathrm{SP}_{\mathrm{depth}}
        =
        \frac{
            \sum_{t=1}^{T}
            d_t^{\mathrm{SP}}
            \mathbb{I}(d_t^{\mathrm{SP}}>\tau_p)
        }{
            \sum_{t=1}^{T}
            \mathbb{I}(d_t^{\mathrm{SP}}>\tau_p)
        }.
        \label{eq:self_penetration}
    \end{equation}

    \item \textbf{Foot Skating.}
    Foot skating measures the residual motion of a foot that should be planted.
    Contact is not read from the evaluated motion but from the source reference,
    so that a motion cannot improve the metric merely by losing contact.
    Let $\mathcal{F}$ contain the two foot sites of the robot model and let
    $\mathbf{p}_{t,f}^{xy}$ be the horizontal position of source foot $f$ at
    source frame $t$.
    The contact indicator is
    \begin{equation}
        c_{t,f}
        =
        \mathbb{I}
        \left(
            \left\|
                \mathbf{p}_{t,f}^{xy}-\mathbf{p}_{t-1,f}^{xy}
            \right\|_2
            \cdot f_s
            <0.01\,\mathrm{m}/\mathrm{s}
        \right),
        \qquad
        f_s=50\,\mathrm{Hz},
    \end{equation}
    following the stance rule of OmniRetarget, which thresholds the horizontal
    source-foot speed and applies no additional height test.
    The label of the interval $(t-1,t)$ belongs to its endpoint frame $t$, and
    the resulting mask is reused unchanged for the dataset, reference, and
    executed motion.
    Foot skating is the mean 3D speed of the evaluated foot over the labeled
    intervals,
    \begin{equation}
        E_{\mathrm{skate}}
        =
        \frac{
            \sum_{t=1}^{T-1}
            \sum_{f\in\mathcal{F}}
            c_{t,f}
            \left\|
                \mathbf{p}_{t,f}-\mathbf{p}_{t-1,f}
            \right\|_2
            \cdot f_s
        }{
            \sum_{t=1}^{T-1}
            \sum_{f\in\mathcal{F}}
            c_{t,f}
        },
        \label{eq:foot_skating}
    \end{equation}
    reported in $\mathrm{cm}/\mathrm{s}$.

    \item \textbf{Foot Floating.}
    Foot floating measures how far a foot that should be planted stays above
    the ground.
    Let $s_{t,f}$ be the smallest signed distance from the floor to the seven
    collision capsules of foot $f$ at frame $t$.
    The clearance is truncated below at zero, because penetration is measured
    separately by the ground-penetration metric:
    \begin{equation}
        E_{\mathrm{float}}
        =
        \frac{
            \sum_{t=1}^{T-1}
            \sum_{f\in\mathcal{F}}
            c_{t,f}
            \left[s_{t,f}\right]_+
        }{
            \sum_{t=1}^{T-1}
            \sum_{f\in\mathcal{F}}
            c_{t,f}
        },
        \label{eq:foot_floating}
    \end{equation}
    evaluated at the labeled endpoint frames and reported in centimeters.
\end{itemize}

\paragraph{Goal Reaching and Execution.}
The goal-reaching experiments evaluate the physically executed motion against
the goal state specified by the condition and against the execution outcome.
Unless stated otherwise, every error below is evaluated at the final frame of
the rollout, and metrics are computed only for the components that the
corresponding condition enables.

\begin{itemize}
    \item \textbf{Goal Position Error ($E_{\mathrm{pos}}$).}
    The position error is the Euclidean distance between the final executed
    root position and the goal root position,
    \begin{equation}
        E_{\mathrm{pos}}
        =
        \left\|
            \mathbf{p}_T-\mathbf{p}_g
        \right\|_2 .
        \label{eq:goal_position_error}
    \end{equation}
    When the goal specifies only a horizontal position, as in the benchmark and
    perturbation protocols, the error is measured in the $xy$ plane and the
    root height is intentionally excluded;
    when the goal specifies a full 3D root position, all three components are
    used.

    \item \textbf{Goal Orientation Error ($E_{\mathrm{ori}}$).}
    The orientation error is the geodesic angle between the final executed root
    orientation and the goal root orientation,
    \begin{equation}
        E_{\mathrm{ori}}
        =
        2\arccos
        \left(
            \left|
                \mathbf{q}_T\cdot\mathbf{q}_g
            \right|
        \right),
        \label{eq:goal_orientation_error}
    \end{equation}
    where $\mathbf{q}_T$ and $\mathbf{q}_g$ are unit quaternions;
    the error is reported in radians.

    \item \textbf{Goal Joint-Configuration Error
    ($E_{\mathrm{MPJPE}}^{\mathrm{goal}}$).}
    The joint-configuration error is the root-relative MPJPE between the body
    positions obtained by forward kinematics from the final executed state and
    from the goal state:
    \begin{equation}
        E_{\mathrm{MPJPE}}^{\mathrm{goal}}
        =
        \frac{1}{J}
        \sum_{j=1}^{J}
        \left\|
            \left(\mathbf{p}_T[j]-\mathbf{p}_T[0]\right)
            -
            \left(\mathbf{p}_g[j]-\mathbf{p}_g[0]\right)
        \right\|_2 ,
        \label{eq:goal_joint_error}
    \end{equation}
    where index $0$ is the root link and $J$ is the number of links.
    Removing the root translation isolates the joint configuration from the
    remaining root-position offset.

    \item \textbf{Goal Root-Velocity Error ($E_{\mathrm{vel}}$).}
    The root-velocity error compares the velocity of the final executed frame,
    obtained by finite differencing the last two frames, with the goal root
    velocity:
    \begin{equation}
        E_{\mathrm{vel}}
        =
        \left\|
            f_s
            \left(
                \mathbf{p}_T-\mathbf{p}_{T-1}
            \right)
            -
            \mathbf{v}_g
        \right\|_2 ,
        \label{eq:goal_velocity_error}
    \end{equation}
    reported in $\mathrm{m}/\mathrm{s}$.

    \item \textbf{Goal End-Effector Error ($E_{\mathrm{ee}}$).}
    The end-effector error is the mean position error over the four
    end-effector anchors, that is, both hands and both feet:
    \begin{equation}
        E_{\mathrm{ee}}
        =
        \frac{1}{4}
        \sum_{e=1}^{4}
        \left\|
            \mathbf{p}_T^{\mathrm{ee},e}
            -
            \mathbf{p}_g^{\mathrm{ee},e}
        \right\|_2 ,
        \label{eq:goal_ee_error}
    \end{equation}
    where the executed and the goal anchors are resolved by the same forward
    kinematics.

    \item \textbf{Fall Rate.}
    Let $h_t$ denote the executed pelvis height.
    An execution is considered fallen if its pelvis height drops below the
    threshold $h_{\min}=0.3\,\mathrm{m}$ at any frame,
    \begin{equation}
        F_i
        =
        \mathbb{I}
        \left(
            \min_{1\le t\le T_i}
            h_{i,t}
            <h_{\min}
        \right).
        \label{eq:fall_indicator}
    \end{equation}
    The fall rate over $N$ executions is
    \begin{equation}
        \mathrm{Fall}
        =
        \frac{100}{N}
        \sum_{i=1}^{N} F_i .
        \label{eq:fall_rate}
    \end{equation}
    Fall is scanned over the complete executed rollout rather than the 2-second
    evaluation window, so a fall that occurs after the window is still counted;
    under the perturbation protocol the scan starts at the force onset.

    \item \textbf{Recovery Rate.}
    The recovery experiment applies a horizontal force of $500\,\mathrm{N}$ for
    $0.1\,\mathrm{s}$ to the torso in a uniformly random direction and then
    commands a standing goal at the current root position.
    Let $\theta_t$ denote the root tilt, that is, the angle between the root
    $+z$ axis and the world vertical direction, which is invariant to yaw.
    A trial counts as fallen when, starting at the force onset, the executed
    motion satisfies both $h_t<0.45\,\mathrm{m}$ and
    $\theta_t>1.0\,\mathrm{rad}$ continuously for $0.2\,\mathrm{s}$, which
    verifies that the perturbation actually caused a fall.
    A fallen trial is recovered when the upright predicate
    \begin{equation}
        U_t
        =
        \mathbb{I}
        \left(
            h_t\ge0.63\,\mathrm{m}
            \;\land\;
            \theta_t\le0.35\,\mathrm{rad}
        \right)
        \label{eq:upright_predicate}
    \end{equation}
    holds continuously for $1.0\,\mathrm{s}$ and the pelvis height never drops
    below $0.3\,\mathrm{m}$ from that moment until the end of the rollout;
    the second condition prevents a trial that stands up and falls again from
    being counted as a recovery.
    The recovery rate is the percentage of fallen trials that recover,
    \begin{equation}
        \mathrm{Recovery}
        =
        \frac{100}{N_{\mathrm{fall}}}
        \sum_{i=1}^{N_{\mathrm{fall}}} R_i .
        \label{eq:recovery_rate}
    \end{equation}

    \item \textbf{Get-Up Time ($T_{\mathrm{getup}}$).}
    For a recovered trial, the get-up time is measured from the force onset to
    the start of the first stable upright interval,
    \begin{equation}
        T_{\mathrm{getup}}
        =
        \frac{k^{\star}}{f_s}-\tau_{\mathrm{on}},
        \label{eq:getup_time}
    \end{equation}
    where $k^{\star}$ is the first frame at which $U_t$ holds continuously for
    $1.0\,\mathrm{s}$ and $\tau_{\mathrm{on}}$ is the force onset time.
    The start of the interval rather than its end is used, because the
    subsequent second validates that the robot was already stably upright at
    that moment.
    $T_{\mathrm{getup}}$ is reported in seconds and averaged over recovered
    trials only, so it must be read jointly with the recovery rate: a method
    that recovers only the easiest trials would otherwise appear
    favorable.
\end{itemize}

\paragraph{Aggregation.}
Each metric is computed per case, averaged within a rollout run, and reported
as the mean and population standard deviation over runs, so the tabulated
spread reflects rollout-seed variation rather than variation in case
difficulty.
The goal-conditioning experiment is the exception: its runs execute the same
trajectory synchronously and deterministically, so its spread is measured
across cases and averaged over runs.
The feature-space metrics are cohort statistics rather than per-case values:
FID, DIV, R@K, and MM-Dist are computed once per run over all cases of that run
and then aggregated across runs, whereas MMod is first averaged within each text
condition and then across conditions.
All reported values are rounded to three decimals.

\end{document}